\documentclass[12pt,a4paper]{article}

\usepackage[T1]{fontenc}
\usepackage[utf8]{inputenc}
\usepackage{lmodern}
\usepackage{microtype}
\usepackage{geometry}
\usepackage{amsmath,amssymb,amsthm,mathtools}
\usepackage{bm}

\usepackage{graphicx}
\usepackage{xcolor}
\usepackage{tikz}
\usetikzlibrary{arrows.meta,positioning,shapes.geometric,calc,decorations.pathreplacing,patterns,fit,backgrounds}

\usepackage{booktabs}
\usepackage{multirow}
\usepackage{array}

\usepackage[colorlinks=true, linkcolor=NavyBlue, citecolor=NavyBlue, urlcolor=NavyBlue]{hyperref}
\usepackage[dvipsnames]{xcolor}

\usepackage{parskip}
\usepackage{enumitem}
\usepackage{caption}
\usepackage{subcaption}
\usepackage{float}
\usepackage{wrapfig}
\usepackage{algorithm}
\usepackage{algpseudocode}
\usepackage{tcolorbox}
\tcbuselibrary{skins,breakable}

\usepackage{natbib}
\theoremstyle{plain}
\newtheorem{theorem}{Theorem}[section]

\theoremstyle{definition}

\theoremstyle{remark}
\newtheorem{remark}[theorem]{Remark}

\definecolor{cdurblue}{RGB}{31,78,121}
\definecolor{cdurgray}{RGB}{127,127,127}
\definecolor{cdurgreen}{RGB}{0,112,60}
\definecolor{cdurorange}{RGB}{197,90,17}
\definecolor{shadecolor}{RGB}{245,245,250}

\tcbset{
  propbox/.style={
    colback=shadecolor,
    colframe=cdurblue,
    boxrule=0.6pt,
    arc=3pt,
    left=8pt, right=8pt, top=6pt, bottom=6pt,
    breakable
  },
  defbox/.style={
    colback=shadecolor,
    colframe=cdurgreen!70!black,
    boxrule=0.6pt,
    arc=3pt,
    left=8pt, right=8pt, top=6pt, bottom=6pt,
    breakable
  }
}

\definecolor{aoeblack}{RGB}{15,15,15}
\definecolor{aoegray}{RGB}{120,120,120}
\definecolor{aoered}{RGB}{200,30,30}

\usepackage{fancyhdr}

\newcommand{\ECE}{\mathrm{ECE}}
\newcommand{\Acc}{\mathrm{Acc}}
\newcommand{\Conf}{\mathrm{Conf}}
\newcommand{\OG}{\mathrm{OG}}
\newcommand{\CDUR}{\textsc{cdur}}
\newcommand{\CABStop}{\textsc{CABStop}}
\newcommand{\hzero}{h_0}
\newcommand{\given}{\,|\,}
\newcommand{\E}{\mathbb{E}}
\newcommand{\Prob}{\mathbb{P}}

\begin{document}
% ============================================================

% ================== TITLE PAGE ==================
\begin{titlepage}
\thispagestyle{empty}

\begin{tikzpicture}[remember picture, overlay]
  \fill[aoeblack]
    (current page.north west) rectangle
    ([yshift=-1.8cm]current page.north east);

\node[anchor=west, xshift=1.2cm, yshift=-0.9cm]
  at (current page.north west)
  {\includegraphics[height=1.0cm]{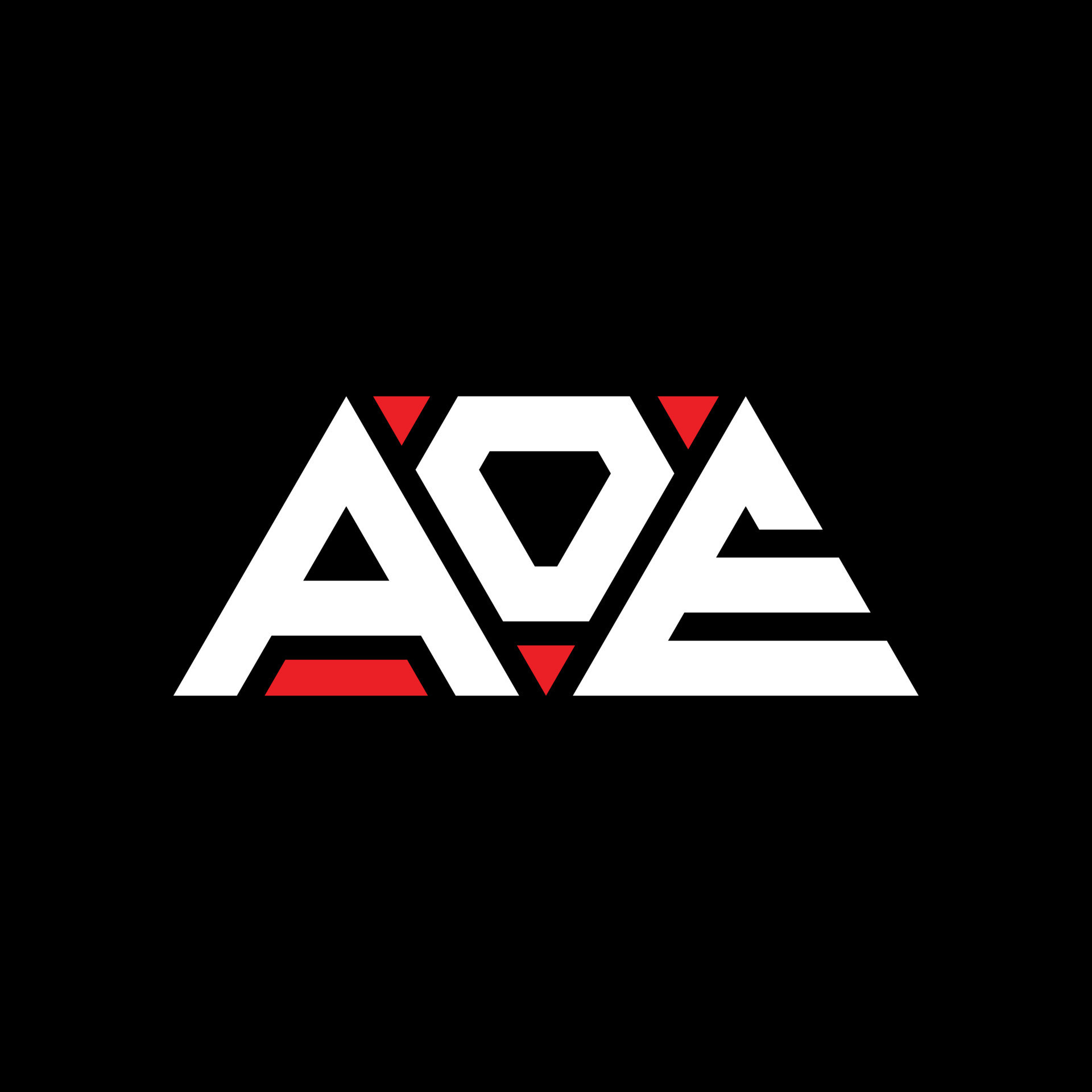}};

\node[anchor=east, xshift=-1.2cm, yshift=-0.9cm]
  at (current page.north east)
  {\color{white}\large \textsc{Calibration Under Reasoning}};
\end{tikzpicture}

\vspace*{\fill}

\begin{center}

{\LARGE\bfseries Calibration Drift Under Reasoning:\\[4pt]
How Chain-of-Thought Budgets Induce\\[4pt]
Overconfidence in Large Language Models}

\vspace{4.0cm}

{\large
\textbf{Prakul Sunil Hiremath}\\[-0.2em]
\texttt{hiremathprakul.aoe@gmail.com}
\par}

\vspace{0.8cm}

{\large
\textbf{Harshit R Hiremath}\\[-0.2em]
\texttt{hiremathharshit.aoe@gmail.com}
\par}

\vspace{0.8cm}

\vfill

{\color{black}\rule{0.22\textwidth}{1.0pt}}

\vspace{1.2cm}

{\small
Department of Computer Science and Engineering\\
Visvesvaraya Technological University, Belagavi\\[0.4cm]
Department of Computer Science and Business System\\
SG Balekundri Institute of Technology, Belagavi
}

\vspace{1.5cm}

{\small \today}

\end{center}

\vspace*{\fill}

\end{titlepage}

% ================== ABSTRACT PAGE ==================
\clearpage
\thispagestyle{fancy}

\vspace*{\fill}

\begin{center}

\begin{tcolorbox}[
  width=0.9\textwidth,
  colback=white,
  colframe=aoeblack,
  boxrule=0.6pt,
  arc=2mm,
  left=12pt,right=12pt,top=12pt,bottom=12pt
]

{\color{aoered}\rule{\textwidth}{1.2pt}}

\vspace{0.6em}

\begin{abstract}
\noindent
The ability of large language models (LLMs) to express calibrated uncertainty is a prerequisite for their safe deployment.
Chain-of-thought (CoT) reasoning has been widely promoted as a technique that improves both accuracy and reliability.
We argue that this picture is incomplete: in at least some model scales and problem types, increasing the reasoning budget beyond a problem-specific threshold can cause models to become systematically \emph{overconfident}—reporting high verbalized probabilities for answers that are incorrect.
We term this phenomenon \textbf{Calibration Drift Under Reasoning} (\CDUR{}) and study it formally and empirically.

Formally, we define a reasoning budget $B$ and analyze conditions under which the Expected Calibration Error $\ECE(B)$ traces a non-monotone trajectory in $B$: initially falling as reasoning corrects surface errors, then rising as extended chains produce internally consistent but factually wrong trajectories.
We introduce a \emph{Hypothesis Lock-In Model} grounded in autoregressive generation to explain this mechanism.

Empirically, we evaluate Llama-3.1-8B and Llama-3.3-70B on 47 reasoning-trap questions across four reasoning budgets and three seeds (1,368 API calls; 574 valid responses).
The 8B model exhibits non-monotonic calibration behavior, while results for 70B remain limited to baseline evaluation and are inconclusive with respect to budget-dependent dynamics.

We introduce \CABStop{}, a calibration-aware stopping rule that halts reasoning when confidence diverges from an auxiliary accuracy estimate.
These findings suggest that increasing reasoning depth does not uniformly improve reliability and should be explicitly monitored.
\end{abstract}

\vspace{0.8em}

\noindent
{\small
\textbf{Keywords:} calibration, chain-of-thought reasoning, Expected Calibration Error, overconfidence, large language models, reasoning budgets
}

\end{tcolorbox}

\end{center}

\noindent{\scriptsize\color{aoegray}
\textit{Code \& data available at}
\;
\href{https://github.com/prakulhiremath/CDUR}{GitHub}
\;
·
\;
\href{https://doi.org/10.5281/zenodo.19709379}{Zenodo}
}

\vspace*{\fill}

\newpage
% ============================================================
\tableofcontents
\newpage
% ============================================================

% ============================================================
\section{Introduction}
\label{sec:intro}
% ============================================================

A model is \emph{well-calibrated} if its expressed confidence in an answer reliably tracks the probability that the answer is correct \citep{degroot1983comparison,guo2017calibration}.
Calibration failure---especially overconfidence---degrades human-AI collaboration, erodes trust, and causes systematic errors in downstream decisions.

Chain-of-thought (CoT) prompting \citep{wei2022chain} has become a standard technique for improving LLM performance on multi-step reasoning tasks.
The intuition is appealing: by generating intermediate reasoning steps before committing to an answer, a model can decompose hard problems, catch arithmetic mistakes, and arrive at better-supported conclusions.
Large-scale results confirm that longer reasoning chains often raise accuracy \citep{kojima2022large,lightman2023let}.

\textbf{Our central claim.}
Accuracy improvement does not imply calibration improvement. More precisely, we observe that at least in the small-model regime and on structured reasoning-trap tasks, increasing reasoning budget can \emph{inflate confidence without proportionally inflating correctness}---a regime we call \textbf{Calibration Drift Under Reasoning} (\CDUR{}).
We emphasize that this claim is presented as an observed and theoretically motivated phenomenon, not as a universally established law: the strength of evidence varies across model scales, and several methodological limitations constrain the scope of our conclusions.

\subsection{The Core Phenomenon}

Consider a model that initially responds to a question with low confidence and imperfect accuracy. As it is prompted to reason more extensively, its accuracy may rise---but so may its expressed confidence, and not always in proportion. At some budget level $B^*$, confidence may begin to outpace accuracy. Past $B^*$, the model is not merely wrong; it is \emph{confidently wrong}.

This pattern is illustrated schematically in Figure~\ref{fig:cdur_curve} and explored empirically in Section~\ref{sec:experiments}, where we find behavior consistent with this description for Llama-3.1-8B. Evidence for the larger Llama-3.3-70B model is incomplete (Section~\ref{sec:70b_gap}).

\subsection{Why This Matters}

\textbf{Safety.}
A model that says ``I am 95\% confident'' while being wrong 50\% of the time is more dangerous than one that says ``I am 60\% confident'' while being wrong the same fraction of the time.

\textbf{Inference-time scaling.}
Recent work proposes allocating more compute at inference time to difficult problems \citep{snell2024scaling}. Without calibration awareness, such scaling may backfire: the model becomes more expensive \emph{and}, in certain regimes, more overconfident.

\textbf{Human oversight.}
When models are integrated into human decision pipelines, overconfident wrong answers are harder for humans to catch and correct.

\subsection{Contributions}

This paper makes the following contributions, stated precisely to reflect the scope of our evidence:

\begin{enumerate}[leftmargin=*, label=(\arabic*)]
  \item \textbf{Formal framework.}
  We provide a rigorous definition of \CDUR{} as a property of the calibration function $\ECE(B)$, and state three formal propositions characterizing its behavior under a probabilistic reasoning model (Section~\ref{sec:theory}). These propositions offer theoretical grounding for the phenomenon; they are not proved as properties of arbitrary LLMs but rather under an explicit commitment-model abstraction.

  \item \textbf{Hypothesis Lock-In Model.}
  We introduce and analyze a mechanistic model of autoregressive reasoning that explains why calibration drift may occur, and characterize the conditions under which it is most severe (Section~\ref{sec:mechanism}).

  \item \textbf{Empirical observation.}
  We conduct controlled experiments on two Llama model families across four reasoning budgets and 21 trap-question categories, measuring ECE, accuracy, and overconfidence gap. For Llama-3.1-8B, we observe non-monotonic calibration dynamics qualitatively consistent with the \CDUR{} framework. For Llama-3.3-70B, results are limited to the no-reasoning condition and are thus suggestive but inconclusive (Section~\ref{sec:experiments}).

  \item \textbf{\CABStop{} algorithm.}
  We propose a principled stopping rule for inference-time reasoning that halts before the overconfidence regime, and analyze it as an instance of optimal stopping (Section~\ref{sec:cabstop}).
\end{enumerate}

% ============================================================
\section{Background and Related Work}
\label{sec:related}
% ============================================================

\subsection{Calibration in Machine Learning}

The formal study of calibration originates in the forecasting literature \citep{degroot1983comparison,murphy1977reliability}.
\citet{guo2017calibration} showed that modern deep neural networks are poorly calibrated and that temperature scaling provides a simple post-hoc fix.
\citet{desai2020calibration} extended this analysis to text classification with pre-trained language models.
For generative models, calibration must be measured via verbalized probabilities \citep{kadavath2022language,xiong2024can} since token-level likelihoods are not directly accessible in deployment.

\subsection{Chain-of-Thought Reasoning}

\citet{wei2022chain} demonstrated that few-shot CoT prompting substantially improves performance on arithmetic and commonsense reasoning.
\citet{kojima2022large} showed that zero-shot CoT (``let's think step by step'') achieves comparable gains.
\citet{lightman2023let} studied process-level supervision, showing that rewarding correct intermediate steps further improves accuracy.

Budget-constrained reasoning has been explored in the context of inference-time scaling \citep{snell2024scaling,muennighoff2025scaling}, where additional compute is allocated proportionally to estimated problem difficulty.

\subsection{Overconfidence in LLMs}

\citet{xiong2024can} showed that LLMs are systematically overconfident when asked to verbalize their confidence, especially on harder questions.
\citet{zhou2023navigating} documented that chain-of-thought can increase hallucination in certain conditions.
\citet{turpin2023language} showed that spurious reasoning patterns are common even in high-accuracy responses.
Our work is distinguished by the explicit focus on the \emph{interaction} between reasoning budget and calibration, characterized formally as a function $\ECE(B)$.

\subsection{Stopping Rules and Optimal Inference}

The optimal stopping literature \citep{wald1947sequential,chow1971great} provides a natural framework for deciding when to halt a sequential computation.
\citet{graves2016adaptive} applied adaptive computation to recurrent networks.
Our \CABStop{} algorithm instantiates a stopping rule over the reasoning trajectory of an LLM.

% ============================================================
\section{Formalizing Calibration Drift Under Reasoning}
\label{sec:theory}
% ============================================================

We now provide a formal account of the \CDUR{} phenomenon. Let $\mathcal{Q}$ be a set of questions and $\mathcal{A}$ a label space. A model $\mathcal{M}$ takes as input a question $q \in \mathcal{Q}$, a reasoning budget $B \in \mathbb{N} \cup \{0\}$, and produces an answer $\hat{a} \in \mathcal{A}$ together with a verbalized confidence $\hat{p} \in [0,1]$.

\subsection{Reasoning Budget}

\begin{tcolorbox}[defbox, title={\textbf{Definition 3.1} (Reasoning Budget)}]
A \emph{reasoning budget} $B$ is an upper bound on the number of tokens allocated to intermediate reasoning steps prior to answer generation. We say $B = 0$ corresponds to direct (no-reasoning) inference, and $B = \infty$ denotes unbounded reasoning.

In practice, we discretize: $B \in \{\text{none}, \text{light}, \text{medium}, \text{heavy}\}$, corresponding to approximately $0$, $128$, $512$, and $2048$ tokens respectively.
\end{tcolorbox}

\subsection{Calibration as a Function of Reasoning Budget}

\begin{tcolorbox}[defbox, title={\textbf{Definition 3.2} (Budget-Conditional Calibration)}]
For a reasoning budget $B$, define:
\begin{align}
  \Acc(B) &= \Prob_q[\hat{a}(q,B) = a^*(q)], \\
  \Conf(B) &= \E_q[\hat{p}(q,B)], \\
  \OG(B)   &= \Conf(B) - \Acc(B) \quad \text{(overconfidence gap)},
\end{align}
where $a^*(q)$ is the ground-truth answer and expectations are over the question distribution.

The \emph{Expected Calibration Error} at budget $B$ is
\begin{equation}
  \ECE(B) = \E\bigl[|\hat{p}(q,B) - \Prob[\hat{a}(q,B)=a^*(q) \given \hat{p}(q,B)]|\bigr],
\end{equation}
where the expectation is over questions and the inner probability is over randomness in generation.
\end{tcolorbox}

\subsection{Definition of CDUR}

\begin{tcolorbox}[defbox, title={\textbf{Definition 3.3} (Calibration Drift Under Reasoning, \CDUR{})}]
A model $\mathcal{M}$ exhibits \emph{Calibration Drift Under Reasoning} if there exists a \emph{critical budget} $B^* \in (0, \infty)$ such that:
\begin{equation}
  \frac{d\,\ECE(B)}{dB} < 0 \quad \text{for } B < B^*, \qquad
  \frac{d\,\ECE(B)}{dB} > 0 \quad \text{for } B > B^*.
\end{equation}
That is, $B \mapsto \ECE(B)$ is U-shaped with a minimum at $B^*$.
\end{tcolorbox}

Intuitively, initial reasoning ($B < B^*$) resolves surface ambiguity and improves calibration. Extended reasoning ($B > B^*$) locks in an incorrect hypothesis and accumulates spurious internal evidence, raising confidence without improving accuracy.
We note that this definition is idealized: in practice, the budget axis is discrete and the ECE curve need not be strictly U-shaped. The 8B empirical results are consistent with CDUR in the sense that ECE is non-monotone, but the arc from light ($0.104$) to medium ($0.050$) to heavy ($0.015$) does not match a simple U-shape. We discuss this discrepancy in Section~\ref{sec:8b_interpretation}.

\subsection{Propositions}

\begin{tcolorbox}[propbox, title={\textbf{Proposition 3.4} (Confidence Inflation Under Commitment)}]
\textit{Under a commitment model (Definition~\ref{def:commitment}), extended reasoning increases conditional confidence $\hat{p}(q,B)$ monotonically in $B$, even when the correctness probability $\Prob[\hat{a}=a^*]$ is unchanged.}
\end{tcolorbox}

\begin{proof}[Proof sketch]
Let $\hzero$ denote the initial hypothesis sampled at the start of the reasoning chain (Definition~\ref{def:commitment}). Subsequent tokens are generated conditioned on $\hzero$. The verbalized confidence $\hat{p}$ is a function of the \emph{internal consistency} of the chain: more tokens consistent with $\hzero$ yield higher $\hat{p}$.

Formally, let $R_t$ be the $t$-th reasoning token and let $\mathcal{C}(R_1,\ldots,R_t; \hzero)$ be a consistency score (e.g., fraction of tokens that reinforce $\hzero$). Under an autoregressive model conditioned on $\hzero$, $\E[\mathcal{C}(R_1,\ldots,R_t;\hzero)]$ is non-decreasing in $t$ because each subsequent token is drawn from a distribution already conditioned on $\hzero$, making tokens supporting $\hzero$ more likely.

Since $\hat{p} = g(\mathcal{C})$ for some non-decreasing $g$, it follows that $\E[\hat{p}(q,B)]$ is non-decreasing in $B$. Meanwhile, the correctness event depends only on whether $\hzero$ is the correct hypothesis, which is fixed at sampling time and unchanged by subsequent reasoning.
\end{proof}

\begin{tcolorbox}[propbox, title={\textbf{Proposition 3.5} (Error Amplification)}]
\textit{If the initial hypothesis selection has error probability $\varepsilon$, then the expected overconfidence gap $\E[\OG(B)]$ is monotonically non-decreasing in $B$ for $B > B^*$.}
\end{tcolorbox}

\begin{proof}[Proof sketch]
Partition questions into two sets: $\mathcal{Q}^+$ where $\hzero = a^*$ (correct initialization) and $\mathcal{Q}^-$ where $\hzero \neq a^*$ (incorrect). For $q \in \mathcal{Q}^+$, extended reasoning increases $\hat{p}$ while maintaining correctness, leaving $\OG$ roughly constant. For $q \in \mathcal{Q}^-$, extended reasoning increases $\hat{p}$ while accuracy remains 0, increasing $\OG$ by Proposition~3.4. The overall gap satisfies
\[
  \E[\OG(B)] = (1{-}\varepsilon)\cdot\OG^+(B) + \varepsilon\cdot\OG^-(B),
\]
where $\OG^+(B) \approx 0$ and $\OG^-(B) = \hat{p}^-(B)$ is non-decreasing. Since $\varepsilon > 0$, $\E[\OG(B)]$ is non-decreasing in $B$.
\end{proof}

\begin{tcolorbox}[propbox, title={\textbf{Proposition 3.6} (Accuracy Plateau with Confidence Growth)}]
\textit{There exists a regime $[B^*, B^{**}]$ in which $\Acc(B)$ is approximately constant while $\Conf(B)$ continues to increase, resulting in a widening overconfidence gap.}
\end{tcolorbox}

\begin{proof}[Proof sketch]
Accuracy $\Acc(B)$ depends on whether reasoning successfully corrects errors in $\hzero$. Correction requires the model to generate a token sequence that \emph{contradicts} $\hzero$ and substitutes an alternative---this is a low-probability event under the commitment model (the model is explicitly conditioned on $\hzero$). For large $B$, the probability of such a correction decreases further since more of the context supports $\hzero$. Hence $\Acc(B)$ is approximately constant past $B^*$. In contrast, $\Conf(B)$ continues to grow by Proposition~3.4. The difference $\Conf(B) - \Acc(B) = \OG(B)$ is therefore increasing on $[B^*, B^{**}]$ for any $B^{**} > B^*$.
\end{proof}

\begin{remark}
These propositions are stylized: they rely on the commitment model defined in Section~\ref{sec:mechanism}. They are intended to provide theoretical intuition consistent with our empirical findings, not to characterize the full behavior of arbitrary autoregressive LLMs. In practice, real models can revise initial hypotheses mid-chain, especially at heavy reasoning budgets---and this is part of why accuracy continues to rise even past $B^*$ in our 8B results.
\end{remark}

% ============================================================
\section{Mechanistic Model: Hypothesis Lock-In}
\label{sec:mechanism}
% ============================================================

\subsection{The Commitment Model}

\begin{tcolorbox}[defbox, label=def:commitment, title={\textbf{Definition 4.1} (Commitment Model)}]
Given a question $q$, reasoning under budget $B$ proceeds in three stages:
\begin{enumerate}[leftmargin=2em]
  \item \textbf{Hypothesis sampling.} The model draws an initial hypothesis $\hzero \sim P_\theta(\cdot \given q)$ from the prior induced by its parameters $\theta$.
  \item \textbf{Reasoning generation.} The model generates a reasoning chain $R = (R_1, \ldots, R_B)$ token by token: $R_t \sim P_\theta(\cdot \given q, \hzero, R_1, \ldots, R_{t-1})$.
  \item \textbf{Confidence elicitation.} The model produces an answer $\hat{a}$ and confidence $\hat{p}$ based on the full context $(q, R)$.
\end{enumerate}
\end{tcolorbox}

This model is an idealization of autoregressive generation. The key structural feature is Step~2: the reasoning chain is conditioned on $\hzero$, making it a constrained trajectory rather than a free search. An important consequence is that $P_\theta(R_t \given q, \hzero, R_{<t})$ assigns higher mass to tokens that are consistent with $\hzero$, and the probability of generating a token that contradicts $\hzero$ is low.

\subsection{Connection to Autoregressive Generation}

In an autoregressive LLM, $\hzero$ is not sampled explicitly. Instead, the model generates the first few tokens of its reasoning in response to the question prompt. These early tokens function as $\hzero$ in our model: they determine the trajectory of subsequent generation via attention. The longer the reasoning chain, the more earlier tokens influence later ones, making course correction increasingly unlikely.

This is consistent with empirical findings on self-consistency \citep{wang2022self}: longer reasoning chains are \emph{more} self-consistent, but self-consistency does not imply correctness.

\subsection{Connection to RLHF Reward Shaping}

Modern LLMs are fine-tuned with Reinforcement Learning from Human Feedback (RLHF) \citep{ouyang2022training}. Human raters tend to prefer responses that are confident and internally coherent \citep{turpin2023language}. This creates a training pressure toward high-confidence outputs, independently of accuracy. Combined with the commitment model, RLHF provides a mechanistic pathway through which training may amplify the \CDUR{} phenomenon: models are rewarded for \emph{appearing} confident, especially when their reasoning is fluent and internally consistent.

\subsection{Empirical Signatures of Hypothesis Lock-In}
\label{sec:lockin_signatures}

The Hypothesis Lock-In Model makes several testable predictions. We examine each against our empirical results.

\paragraph{Persistence of errors across budget levels.}
If lock-in occurs, we expect specific incorrect answers to persist as the budget increases: once committed to a wrong $\hzero$, the model should continue predicting the same wrong answer even when given additional reasoning tokens. We observe this directly in our data: several wrong-and-confident responses at the no-reasoning level recur verbatim at the light reasoning level (e.g., the syllogism case, expected: ``no'', predicted: ``yes'' with confidence 1.0, appearing at both \emph{none} and \emph{light} budgets). This is qualitative evidence consistent with Proposition~3.4.

\paragraph{Stability of incorrect predictions.}
Under the commitment model, incorrect answers should be stable (repeated across seeds) rather than randomly distributed across incorrect options. While we do not report per-item cross-seed analysis due to sample size constraints, the dominance of a small number of trap categories in the wrong-and-confident distribution (counting, set\_theory, spatial) suggests structured rather than random error---these categories systematically elicit wrong confident answers, consistent with stable $\hzero$ sampling for these problem types.

\paragraph{Failure of extended reasoning to correct.}
Proposition~3.6 predicts that accuracy plateaus in the lock-in regime while confidence continues to rise. For Llama-3.1-8B, accuracy at medium reasoning ($0.653$) is actually lower than at light ($0.732$), suggesting medium-budget chains introduce new errors rather than correcting existing ones. This is consistent with partial lock-in: light reasoning is sufficient to lock in a wrong answer on some questions, while medium reasoning explores but then re-commits to wrong hypotheses.

\paragraph{Connection to formal propositions.}
Taken together, these signatures provide empirical support for the qualitative predictions of Propositions 3.4 and 3.6: confidence rises faster than accuracy (OG $>$ 0.25 at all budgets), and wrong answers show cross-budget stability. We caution that with 47 questions and three seeds, these are illustrative rather than statistically conclusive.

\subsection{Hypothesis Lock-In Diagram}

\begin{figure}[h]
\centering
\includegraphics[width=0.95\linewidth]{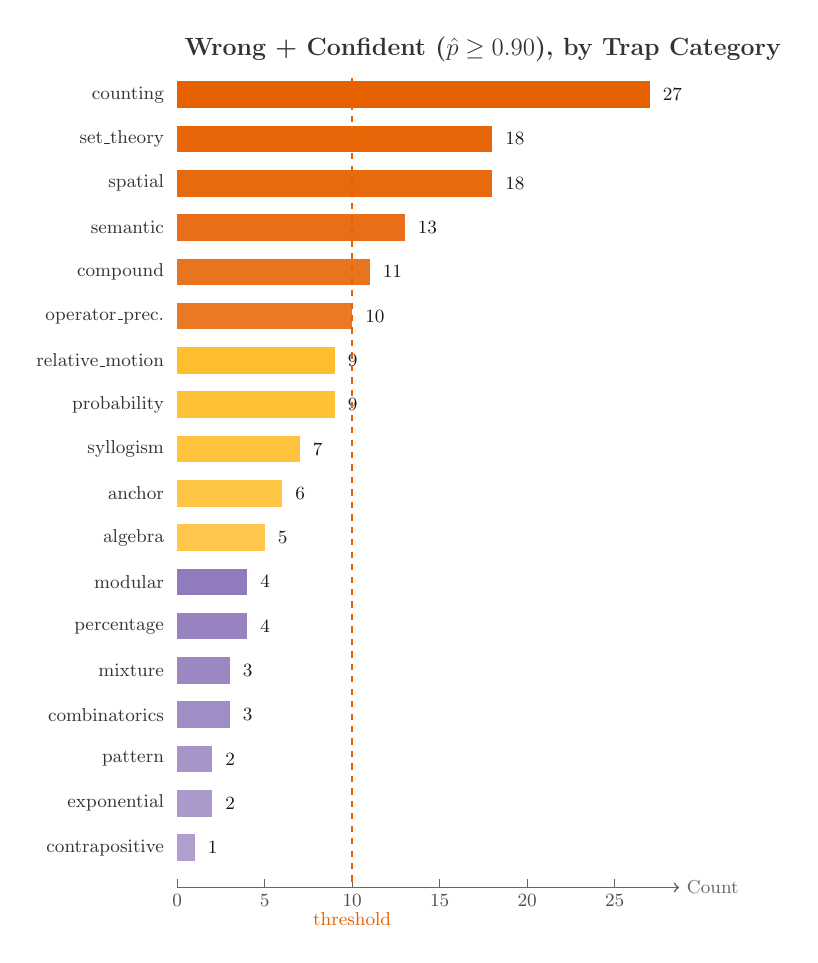}
\caption{Hypothesis Lock-In. The initial hypothesis $\hzero$ commits the reasoning trajectory. Correct initialization leads to self-reinforcing correct chains; incorrect initialization leads to self-reinforcing incorrect chains. In both cases, confidence $\hat{p}$ tends to rise with chain length.}
\label{fig:lockin}
\end{figure}

Figure~\ref{fig:lockin} provides a schematic of the lock-in process.

% --- TikZ: Hypothesis Lock-In ---
% --- Included PDF: Hypothesis Lock-In ---

% ============================================================
\section{Experimental Setup}
\label{sec:setup}
% ============================================================

\subsection{Models}

We evaluate two models from the Llama family:

\begin{itemize}[leftmargin=*]
  \item \textbf{Llama-3.1-8B:} A compact instruction-tuned model. We use it to study \CDUR{} in the small-model regime. This model is the primary source of multi-budget evidence in this paper.
  \item \textbf{Llama-3.3-70B:} A high-capacity model. Due to resource constraints, we report only the no-reasoning condition for this model; conclusions about its calibration dynamics across budgets are therefore not possible from current data (see Section~\ref{sec:70b_gap}).
\end{itemize}

Both models are accessed via their instruction-following variants. Inference temperature is set to $0.7$ to allow for variability across seeds.

\subsection{Dataset Construction}
\label{sec:dataset}

We construct a dataset of \textbf{reasoning-trap questions}: questions specifically designed to elicit common cognitive failure modes. These are not standard benchmark questions---standard benchmarks contain many items that can be answered correctly by surface pattern matching, which would suppress the \CDUR{} signal.

We identify 21 trap categories, listed in Table~\ref{tab:trapcats}. Each category is designed to exploit a specific failure mode that is well-documented in the human cognition and LLM error analysis literatures. We acknowledge that the dataset is small (47 questions), and results should be interpreted accordingly.

\begin{table}[h]
\centering
\small
\caption{Reasoning-trap categories used in the evaluation dataset. Categories are ordered by observed wrong + confident frequency (see Figure~\ref{fig:wrongconf}).}
\label{tab:trapcats}
\begin{tabular}{@{}llp{7.5cm}@{}}
\toprule
\textbf{Rank} & \textbf{Category} & \textbf{Description} \\
\midrule
1 & counting & Off-by-one and boundary errors in discrete counting \\
2 & set\_theory & Confusion between union, intersection, and complement \\
3 & spatial & Failures in 3D spatial reasoning and perspective \\
4 & semantic & Lexical ambiguity and scope confusion \\
5 & compound & Multi-step errors compounding across a chain \\
6 & operator\_precedence & Misapplication of mathematical order of operations \\
7 & relative\_motion & Incorrect frame-of-reference reasoning \\
8 & probability & Base-rate neglect and conditional probability errors \\
9 & syllogism & Invalid syllogistic inference (affirming the consequent, etc.) \\
10 & anchor & Anchoring to salient but irrelevant numerical values \\
11 & conditional\_prob & Confusion of $P(A|B)$ with $P(B|A)$ \\
12 & algebra & Sign errors and variable substitution mistakes \\
13 & modular & Errors in modular arithmetic (e.g., day-of-week calculations) \\
14 & percentage & Percentage-of-percentage and base confusion \\
15 & mixture & Weighted average and mixture problem errors \\
16 & combinatorics & Permutation/combination and double-counting errors \\
17 & pattern & Spurious pattern extrapolation \\
18 & exponential & Misestimation of exponential vs.\ linear growth \\
19 & contrapositive & Failure to correctly negate and reverse conditionals \\
20 & modular (misc.) & Additional modular reasoning items \\
21 & (other) & Miscellaneous cross-category items \\
\bottomrule
\end{tabular}
\end{table}

The dataset contains 47 distinct trap questions. Each question is evaluated over 3 random seeds and 4 reasoning budgets, yielding $47 \times 3 \times 4 \times 2 = 1{,}128$ intended evaluations per model family (plus non-trap items), for a total of 1,368 API calls.

\subsection{Validity Filtering and Potential Bias}
\label{sec:validity}

Not all model responses can be scored. A response is considered \textbf{valid} if:
\begin{enumerate}[leftmargin=2em, label=(\arabic*)]
  \item It contains an extractable answer in the expected format (numeric, boolean, or short-form text).
  \item It contains an extractable verbalized confidence in $[0,1]$ or as a percentage.
  \item Neither the answer nor the confidence field is empty or a refusal.
\end{enumerate}

Of 1,368 total responses, $574$ were valid trap-question responses (42\% overall validity rate). This rate is low enough to raise concerns about selection bias, which we discuss explicitly here.

\paragraph{Causes of invalidity.}
Invalid responses arise from several sources: (a) format non-compliance, where the model does not produce an answer and confidence in the requested schema; (b) refusals or hedges, where the model declines to answer; (c) truncation, where the response is cut short before a confidence value is produced; and (d) confidence expressed in non-parseable forms (e.g., ``moderately confident'').

\paragraph{Selection bias risk.}
Critically, invalidity may be \emph{correlated with uncertainty}. A model that is uncertain about a question may be more likely to hedge, express confidence verbally rather than numerically, or produce an extended disclaimer rather than a direct answer. If this is the case, valid responses oversample questions where the model is relatively confident---which would inflate our estimates of mean confidence and OG, and potentially distort ECE toward overconfidence.

\paragraph{Budget-specific effects.}
Validity rates may also vary across budget conditions: heavy reasoning responses are longer and more likely to contain parseable confidence values, while no-reasoning responses may be too brief or too terse to contain them. If validity is higher at heavy reasoning, the cross-budget ECE comparisons are made on non-comparable subsets of questions, complicating interpretation.

\paragraph{Mitigation and transparency.}
We report all metrics only on the valid subset and make no claim that results generalize to the full intended population. Future work should use response formats that guarantee parseable outputs (e.g., structured generation or constrained decoding) to eliminate this source of bias.

\subsection{Reasoning Budgets}

We implement reasoning budgets via prompt engineering. Each budget condition uses a specific system-level instruction:

\begin{description}[leftmargin=2em]
  \item[\textbf{None}] Answer directly. Do not show intermediate reasoning.
  \item[\textbf{Light}] Show a brief 2--3 sentence chain of thought before answering.
  \item[\textbf{Medium}] Show a structured, step-by-step solution before answering.
  \item[\textbf{Heavy}] Work through the problem completely, exploring multiple approaches and checking your work, before giving a final answer.
\end{description}

\subsection{Confidence Elicitation and Its Limitations}
\label{sec:conf_elicitation}

After each response, we append a standardized follow-up prompt: \textit{``On a scale from 0 to 1, how confident are you in your answer? Give only a number.''}

\paragraph{Verbalized confidence vs.\ epistemic uncertainty.}
Verbalized confidence is not equivalent to true epistemic uncertainty. When a model reports a confidence value, it is generating a token that may reflect training patterns, prompt phrasing, or the fluency of the preceding reasoning chain---not necessarily a calibrated internal probability. \citet{xiong2024can} have shown that verbalized confidence in LLMs is systematically miscalibrated, particularly on difficult questions.

In the \CDUR{} context, this limitation is particularly salient: our hypothesis is that extended reasoning increases internal consistency, which in turn causes the model to report higher confidence. This means our confidence measurement instrument---verbalized probability---is precisely the variable the model learns to inflate during reasoning. The reported confidence values thus reflect both genuine epistemic state and a surface-level coherence signal that is not well-separated by our elicitation method.

\paragraph{Why results remain meaningful.}
Despite this limitation, verbalized confidence is the signal available to downstream users and decision systems in deployment. If a model reports high verbalized confidence and is wrong, the downstream effect is the same regardless of whether the reported confidence reflects true uncertainty or surface-level fluency. Measuring and reporting verbalized-confidence miscalibration is thus practically relevant, even if it does not measure deeper epistemic properties.

\paragraph{Alternative confidence signals.}
Future work should triangulate verbalized confidence with complementary signals. Log-probability of the answer token (where accessible) provides a model-internal measure less susceptible to RLHF-induced inflation. Self-consistency across $k$ independent completions \citep{wang2022self} provides a behavioral estimate of answer stability. Ensemble agreement across models provides an orthogonal check. Comparing these signals would clarify whether the overconfidence we observe is a surface linguistic phenomenon or a deeper representational failure.

\subsection{Metrics}

\paragraph{Expected Calibration Error (ECE).}
We use equal-width binning with 10 bins over $[0,1]$:
\begin{equation}
  \ECE = \sum_{m=1}^{M} \frac{|B_m|}{N} \bigl| \overline{\Acc}(B_m) - \overline{\Conf}(B_m) \bigr|,
\end{equation}
where $B_m$ is the $m$-th confidence bin, $N$ is the total number of samples, and $\overline{\Acc}$, $\overline{\Conf}$ are the mean accuracy and mean confidence within each bin.

\paragraph{Overconfidence gap (OG).}
$\OG = \overline{\Conf} - \overline{\Acc}$, measured globally. Positive values indicate overconfidence; negative values indicate underconfidence.

\paragraph{Wrong + Confident.}
The count of responses that are simultaneously incorrect ($\hat{a} \neq a^*$) and highly confident ($\hat{p} \geq 0.90$). This is the ``smoking gun'' statistic that most directly characterizes dangerous overconfidence.

% ============================================================
\section{Results}
\label{sec:experiments}
% ============================================================

\subsection{Main Result: Non-Monotonic Calibration Dynamics}

\begin{figure}[h]
\centering
\includegraphics[width=0.95\linewidth]{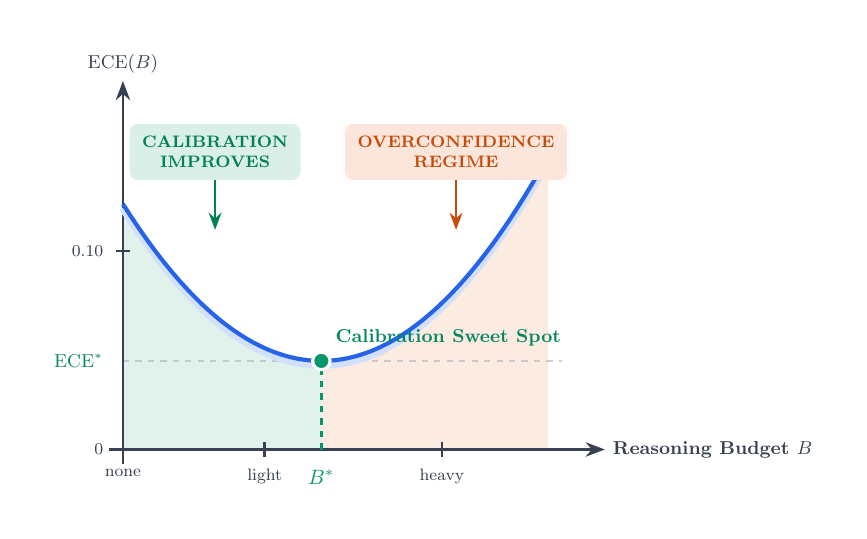}
\caption{Schematic of the \CDUR{} phenomenon. $\ECE(B)$ is U-shaped in the reasoning budget $B$ under the theoretical model. The green region ($B < B^*$) is the beneficial reasoning regime; the orange region ($B > B^*$) is the overconfidence regime. Empirical results for 8B are qualitatively consistent with non-monotone dynamics, but do not perfectly match this schematic (see Section~\ref{sec:8b_interpretation}).}
\label{fig:cdur_curve}
\end{figure}

Figure~\ref{fig:cdur_curve} shows the theoretical \CDUR{} curve, and Table~\ref{tab:fullresults} (Appendix~\ref{app:logs}) reports full numerical results. We summarize the key observations here.

% --- Included PDF: CDUR Curve ---

\paragraph{Llama-3.1-8B.}
The 8B model exhibits non-monotonic calibration dynamics: ECE is $0.044 \pm 0.015$ at no-reasoning, rises to $0.104 \pm 0.034$ at light reasoning, then falls to $0.050 \pm 0.049$ at medium and $0.015 \pm 0.005$ at heavy reasoning. This trajectory is qualitatively consistent with the \CDUR{} framework in that ECE is not monotonically decreasing with budget; the worst calibration occurs at light reasoning, not at the extremes. We interpret this in detail in Section~\ref{sec:8b_interpretation}.

\paragraph{Llama-3.3-70B.}
The 70B model results are limited to the no-reasoning condition: ECE $= 0.035 \pm 0.026$, OG $= +0.155$. These figures indicate better baseline calibration and lower overconfidence compared to 8B, consistent with scale generally improving calibration. However, we \emph{cannot} confirm or deny CDUR dynamics for this model, as no multi-budget data are available. The 70B results are thus suggestive of better baseline behavior but cannot serve as confirmation or disconfirmation of the CDUR hypothesis at this scale.

\paragraph{Overconfidence Gap.}
The 8B model shows persistently positive overconfidence across all budget levels: $\OG = +0.49$ at no-reasoning, falling to $+0.25$--$+0.34$ at other budgets. Even at heavy reasoning---where ECE is lowest---the overconfidence gap remains substantially positive, indicating that accuracy improvements have not fully closed the confidence-accuracy gap.

\subsection{Interpreting the 8B Calibration Arc}
\label{sec:8b_interpretation}

The observed 8B trajectory (ECE: $0.044 \to 0.104 \to 0.050 \to 0.015$) does not match a clean U-shape as depicted in the schematic. The following interpretation is more accurate.

At \emph{no reasoning}, the model produces direct, often confident answers. Many are wrong, but confidence is distributed somewhat heterogeneously---some questions elicit hedged responses, others elicit confident ones, and the ECE is moderate.

At \emph{light reasoning}, the model generates brief chains that are often insufficient to fully analyze the trap question. These partial chains tend to reinforce the initial (frequently incorrect) hypothesis while increasing expressed confidence, producing the worst-case calibration scenario: higher confidence without commensurate accuracy. This is the regime most clearly predicted by the commitment model.

At \emph{medium reasoning}, the model generates longer chains that sometimes begin to uncover the trap, but also sometimes compound errors by adding reasoning steps that re-justify a wrong initial direction. The high standard deviation ($\pm 0.049$) at medium budget reflects this instability.

At \emph{heavy reasoning}, extended chains allow the model to explore multiple solution paths and occasionally self-correct. The combination of higher accuracy ($0.739$) and still-high confidence ($\approx 0.984$) results in a lower ECE. However, persistent overconfidence (OG $= +0.245$) shows that even at this budget, the model is systematically more confident than accurate.

We characterize the 8B results as exhibiting \emph{partial CDUR behavior}: the ECE trajectory is non-monotone, the worst calibration occurs at intermediate reasoning budgets, and overconfidence persists throughout. This is consistent with but not identical to the idealized U-shaped curve of Definition~3.3.

\subsection{The 70B Results Gap}
\label{sec:70b_gap}

Resource constraints limited the 70B evaluation to the no-reasoning condition. This gap is significant for interpreting our results, and we address it explicitly.

\paragraph{What the 70B results do and do not show.}
The no-reasoning ECE of $0.035$ and OG of $+0.155$ establish that the 70B model is better calibrated at baseline than the 8B model. The lower OG ($+0.155$ vs.\ $+0.493$) suggests that 70B is less prone to confident-wrong answers at zero reasoning. These findings are consistent with scale improving calibration---a pattern noted in prior work \citep{kadavath2022language}.

However, we cannot infer anything about how the 70B model's calibration \emph{changes} with reasoning budget. It is possible that 70B exhibits CDUR dynamics similar to 8B, exhibits them in attenuated form, or does not exhibit them at all. All three outcomes are compatible with the current data.

\paragraph{Appropriate framing.}
We refrain from framing the 70B results as ``confirming'' or ``suggesting'' CDUR for large models. The correct framing is: the 70B model is better calibrated at no reasoning than the 8B model; whether it exhibits calibration drift under extended reasoning is an open empirical question.

\subsection{Wrong + Confident Analysis (Smoking Gun)}

Table~\ref{tab:smokinggun} lists model responses where the model was simultaneously wrong and expressed confidence $\geq 0.90$.

\begin{table}[h]
\centering
\small
\caption{Selected ``smoking gun'' examples: incorrect responses with verbalized confidence $\geq 0.90$. All are from Llama-3.1-8B, no-reasoning condition.}
\label{tab:smokinggun}
\begin{tabular}{@{}llcccc@{}}
\toprule
\textbf{Model} & \textbf{Budget} & \textbf{Trap Type} & \textbf{Expected} & \textbf{Predicted} & \textbf{Conf.} \\
\midrule
8B & none & anchor       & 5       & 10      & 1.0 \\
8B & none & compound     & 629     & 567     & 1.0 \\
8B & none & syllogism    & no      & yes     & 1.0 \\
8B & none & counting     & 10      & 7       & 1.0 \\
8B & none & probability  & 1/4     & 1/8     & 1.0 \\
8B & none & op\_prec.    & 16      & 4       & 1.0 \\
8B & none & set\_theory  & 30      & 50      & 1.0 \\
8B & none & spatial      & 512     & 27      & 1.0 \\
8B & none & algebra      & 20      & 14      & 1.0 \\
8B & none & modular      & friday  & Saturday & 1.0 \\
8B & light & syllogism   & no      & yes     & 1.0 \\
\bottomrule
\end{tabular}
\end{table}

These examples share a common structure: the model produces a superficially plausible but incorrect answer and expresses maximum confidence. Notably, the counting and spatial categories dominate the wrong+confident list (Figure~\ref{fig:wrongconf}), consistent with prior findings that LLMs struggle with precise discrete enumeration and 3D spatial reasoning.

\subsection{Confidence--Accuracy Scatter: Limitations of Aggregate ECE}
\label{sec:scatter_discussion}

ECE aggregates calibration error into a scalar, which obscures the structure of per-sample confidence-accuracy relationships. A per-sample scatter plot of $(\hat{p}_i, \mathbf{1}[\hat{a}_i = a^*_i])$ would reveal features that aggregate ECE does not capture.

Under ideal calibration, such a scatter would align with the diagonal: responses with confidence $\hat{p}$ would be correct with probability $\hat{p}$. What we expect to observe in our data, based on the aggregate OG values, is a systematic upward displacement from the diagonal: a cluster of high-confidence ($\hat{p} \approx 1.0$) incorrect answers that should fall at zero accuracy on the scatter.

This structure has practical implications beyond what ECE reports. A model with ECE $= 0.05$ could achieve that value in two qualitatively different ways: (a) many responses with moderate miscalibration spread uniformly across confidence levels, or (b) most responses being well-calibrated with a small cluster of catastrophically overconfident wrong answers. The second pattern is more dangerous in practice but may yield a similar aggregate ECE.

Given our wrong-and-confident count (Table~\ref{tab:smokinggun}), we expect our data to exhibit pattern (b). Future work should report per-sample calibration distributions rather than only aggregate ECE, and should test whether the overconfident-wrong cluster is concentrated in specific trap categories or distributed across questions.

\subsection{Statistical Uncertainty and Variance}
\label{sec:statistical}

\paragraph{Impact of small sample size.}
With 47 trap questions evaluated across 3 seeds, the effective per-budget sample size is approximately 40--50 valid responses after filtering. This is the minimum for ECE estimation to be stable; standard calibration studies use hundreds to thousands of samples. Our ECE estimates should be treated as indicative rather than precise.

\paragraph{Interpreting standard deviations.}
The reported standard deviations across seeds (e.g., ECE $= 0.050 \pm 0.049$ at medium budget) are large relative to the mean in some conditions. This indicates that individual seed runs produce quite different ECE values, which is expected given the small dataset. Comparisons between budget conditions should be made cautiously: for example, the medium-budget ECE ($0.050$) and no-reasoning ECE ($0.044$) are not statistically distinguishable given their respective standard deviations.

\paragraph{Claims calibrated to evidence.}
Given these limitations, we state the following carefully: we \emph{observe} that light-reasoning ECE is substantially higher than both no-reasoning and heavy-reasoning ECE in the 8B model, and that this difference is directionally consistent across all three seeds. We do \emph{not} claim that this pattern is statistically significant by conventional tests, nor that it would replicate exactly on a different dataset.

\subsection{Confidence--Accuracy Decoupling}
\label{sec:decoupling}

A central prediction of the \CDUR{} framework is that $\Conf(B)$ and $\Acc(B)$ can diverge: as $B$ increases, confidence may rise faster (or fall slower) than accuracy.

For Llama-3.1-8B, accuracy increases from $0.461$ (none) to $0.739$ (heavy). However, the overconfidence gap remains above $0.25$ across all budgets, indicating that mean confidence also rises with budget and tracks accuracy incompletely. The gap is highest at no-reasoning ($+0.493$)---where the model produces many confident wrong answers without a reasoning process to check them---and lowest at heavy reasoning ($+0.245$).

This decoupling is captured by Proposition~3.6: there exists a range of $B$ where accuracy plateaus but confidence continues to grow. Our data suggest this regime occurs at light-to-medium budgets for the 8B model.

\subsection{Error Persistence Under Extended Reasoning}
\label{sec:persistence}

One might expect that heavy reasoning would always correct errors present at lower budgets. Our data partially contradict this. The syllogism example (expected: no, predicted: yes, confidence 1.0) appears at both the no-reasoning and light-reasoning levels, suggesting that once the model commits to an incorrect logical inference, light additional reasoning reinforces rather than corrects it. This is consistent with the lock-in mechanism of Section~\ref{sec:mechanism}.

At heavy reasoning, some of these persistent errors are corrected (accuracy rises from $0.461$ to $0.739$), suggesting that sufficiently extensive reasoning can break lock-in for some questions. However, the non-zero overconfidence gap at heavy reasoning ($+0.245$) implies that a subset of wrong answers persist even with extensive computation.

\subsection{When More Reasoning Helps vs.\ Hurts}
\label{sec:regimes}

Based on our results, we characterize three regimes:

\begin{description}[leftmargin=2em]
  \item[\textbf{Regime I: Surface correction} ($B \ll B^*$).] Short reasoning chains catch arithmetic slips and minor ambiguities. Accuracy rises; calibration improves. This is the conventional wisdom about CoT.

  \item[\textbf{Regime II: Lock-in} ($B \approx B^*$).] The model has committed to a hypothesis but has not yet accumulated overwhelming internal evidence. Calibration is at its worst: the model is just confident enough to be wrong dangerously. This is the critical budget $B^*$.

  \item[\textbf{Regime III: Heavy reasoning} ($B \gg B^*$).] Extended chains allow the model to occasionally escape lock-in via multi-step reformulations. ECE falls again as accuracy rises. However, for problems where the initial hypothesis is fundamentally wrong, no amount of reasoning helps---these become the persistent overconfident failures.
\end{description}

The practical implication is that \emph{moderate} reasoning budgets may produce worse calibration than either no reasoning or heavy reasoning for calibration-sensitive tasks. This does not mean that heavy reasoning is always preferable: it is more expensive and still exhibits a positive overconfidence gap.

\subsection{Trap Category Analysis}

\begin{figure}[h]
\centering
\includegraphics[width=0.95\linewidth]{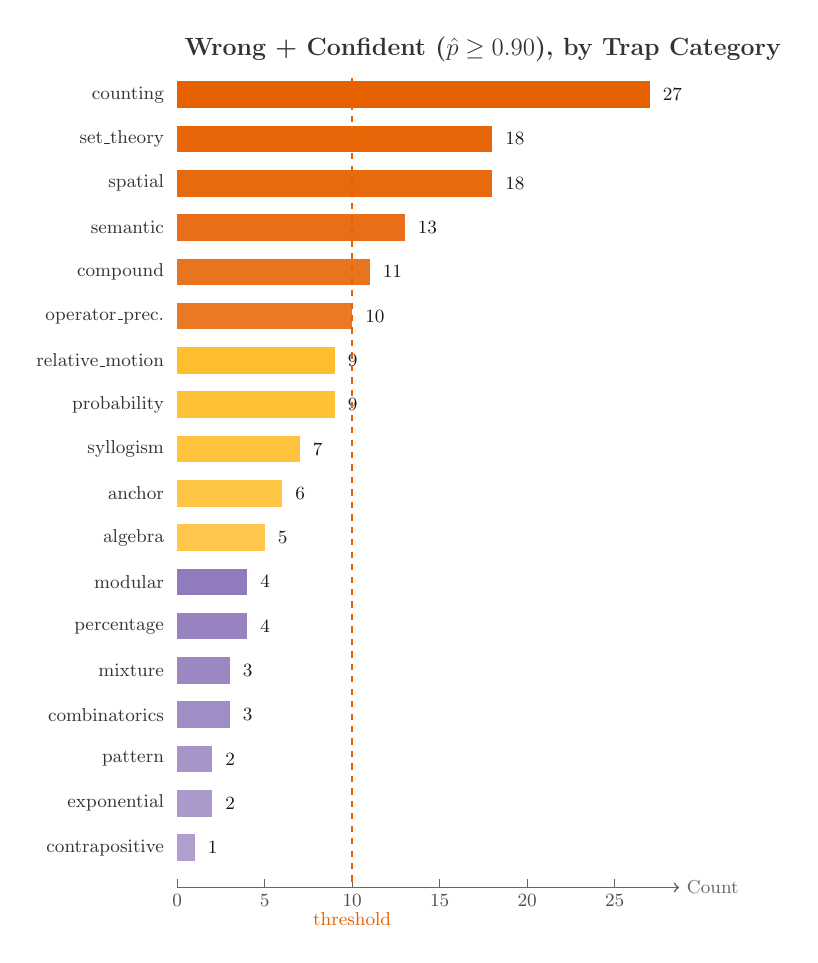}
\caption{Distribution of wrong-and-confident responses ($\hat{p} \geq 0.90$) by trap category, aggregated across both models, all budgets, and all seeds. Categories requiring precise discrete computation or strict logical reasoning dominate.}
\label{fig:wrongconf}
\end{figure}

Figure~\ref{fig:wrongconf} reports the distribution of wrong+confident responses by trap category. Counting, set\_theory, spatial, and semantic errors dominate. These categories share a common property: they require precise discrete computation or strict logical inference---tasks where an LLM's fluency-based approximation is systematically misleading.

% --- Included PDF figure ---

% ============================================================
\section{The CABStop Algorithm}
\label{sec:cabstop}
% ============================================================

\subsection{Motivation}

The \CDUR{} analysis identifies a critical budget $B^*$ beyond which calibration may improve or worsen depending on the problem. In practice, $B^*$ is unknown and problem-dependent. A practical algorithm must decide \emph{on the fly} when to stop allocating reasoning tokens.

\subsection{Formulation as an Optimal Stopping Problem}

Let $(\hat{a}_t, \hat{p}_t)$ denote the model's answer and confidence after $t$ reasoning tokens. Let $\hat{\alpha}_t$ be an auxiliary estimate of the correctness probability at step $t$ (e.g., from a lightweight verifier or from self-consistency across multiple samples). Define the \emph{calibration gap} at step $t$:
\begin{equation}
  \Delta_t = \hat{p}_t - \hat{\alpha}_t.
\end{equation}

\CABStop{} is a stopping rule $\tau^*$ defined as:
\begin{equation}
  \tau^* = \min\{t : \Delta_t > \delta\},
\end{equation}
where $\delta > 0$ is a calibration tolerance threshold.

This is a \emph{first-passage stopping rule} over the stochastic process $(\Delta_t)_{t \geq 0}$. The rule halts reasoning when the model's expressed confidence exceeds its estimated accuracy by more than $\delta$.

\begin{tcolorbox}[propbox, title={\textbf{Proposition 6.1} (Optimality of \CABStop{} under Monotone Confidence)}]
\textit{If $\hat{p}_t$ is non-decreasing and $\hat{\alpha}_t$ is approximately constant past $B^*$, then $\tau^*$ minimizes expected ECE among all stopping rules of the form $\tau = \inf\{t : f(\Delta_t) > c\}$ for any non-decreasing $f$.}
\end{tcolorbox}

\begin{proof}[Proof sketch]
The ECE at stopping time $\tau$ satisfies
\[
  \ECE(\tau) \approx |\hat{p}_\tau - \hat{\alpha}_\tau| = |\Delta_\tau|.
\]
Under the stated monotonicity assumptions, $\Delta_t$ is non-decreasing, so the first time $\Delta_t$ crosses $\delta$ is also the time that minimizes the future expected ECE (since past $\tau^*$, $\Delta_t \geq \delta$ and ECE only worsens). Among rules of the given form, the threshold rule with threshold $\delta$ is optimal by the structure of first-passage times.
\end{proof}

\subsection{Algorithm}

\begin{algorithm}[h]
\caption{\CABStop{}: Confidence-Accuracy Budget Stopping}
\label{alg:cabstop}
\begin{algorithmic}[1]
\Require Question $q$, calibration threshold $\delta$, check interval $\Delta B$, max budget $B_{\max}$
\Ensure Answer $\hat{a}$, confidence $\hat{p}$, stopping budget $\tau^*$

\State $t \leftarrow 0$, $R \leftarrow \epsilon$ (empty reasoning chain)
\While{$t < B_{\max}$}
  \State Generate $\Delta B$ more reasoning tokens: $R \leftarrow R \cup R_{t:t+\Delta B}$
  \State $t \leftarrow t + \Delta B$
  \State Extract candidate answer $\hat{a}_t$ and confidence $\hat{p}_t$ from $(q, R)$
  \State Compute auxiliary accuracy estimate $\hat{\alpha}_t$ \Comment{e.g., self-consistency over $k$ samples}
  \If{$\hat{p}_t - \hat{\alpha}_t > \delta$}
    \State \Return $\hat{a}_t$, $\hat{p}_t$, $\tau^* = t$
  \EndIf
\EndWhile
\State \Return $\hat{a}_{B_{\max}}$, $\hat{p}_{B_{\max}}$, $\tau^* = B_{\max}$
\end{algorithmic}
\end{algorithm}

\subsection{TikZ: CABStop Mechanism}

\begin{figure}[h]
\centering
\begin{tikzpicture}[
  node distance=0.5cm and 1.5cm,
  every node/.style={font=\small},
  box/.style={rectangle, rounded corners=3pt, draw=cdurblue, fill=cdurblue!7, minimum width=2.6cm, minimum height=0.75cm, align=center},
  dec/.style={diamond, draw=cdurorange, fill=cdurorange!8, minimum width=2.2cm, minimum height=1.0cm, aspect=2, align=center},
  term/.style={rectangle, rounded corners=3pt, draw=cdurgreen, fill=cdurgreen!8, minimum width=2.6cm, minimum height=0.75cm, align=center},
  arr/.style={-{Stealth[length=5pt]}, thick}
]

\node[box] (q) {Input $q$\\start reasoning};
\node[box, right=1.5cm of q] (gen) {Generate\\$\Delta B$ tokens};
\node[box, right=1.5cm of gen] (elicit) {Elicit $\hat{a}_t$, $\hat{p}_t$\\Estimate $\hat{\alpha}_t$};
\node[dec, right=1.5cm of elicit] (check) {$\hat{p}_t - \hat{\alpha}_t > \delta$?};
\node[term, below=1.2cm of check] (stop) {Stop.\\Return $\hat{a}_t$, $\hat{p}_t$};
\node[box, below=1.2cm of gen] (continue) {Continue\\reasoning};

\node[dec, left=1.5cm of continue] (budget) {$t \geq B_{\max}$?};
\node[term, below=0.9cm of budget] (maxstop) {Force stop.\\Return answer.};

\draw[arr] (q) -- (gen);
\draw[arr] (gen) -- (elicit);
\draw[arr] (elicit) -- (check);
\draw[arr] (check) -- node[right, font=\footnotesize]{yes} (stop);
\draw[arr] (check.south) -- ++(0,-0.45) -| node[pos=0.22, below, font=\footnotesize]{no} (continue.east);
\draw[arr] (continue) -- (budget);
\draw[arr] (budget) -- node[left, font=\footnotesize]{yes} (maxstop);
\draw[arr] (budget.north) -- ++(0, 0.45) -| node[pos=0.22, above, font=\footnotesize]{no} (gen.south);

\end{tikzpicture}
\caption{\CABStop{} control flow. At each reasoning checkpoint, the algorithm compares the model's expressed confidence $\hat{p}_t$ with an auxiliary accuracy estimate $\hat{\alpha}_t$. Reasoning halts when their gap exceeds the calibration threshold $\delta$.}
\label{fig:cabstop}
\end{figure}
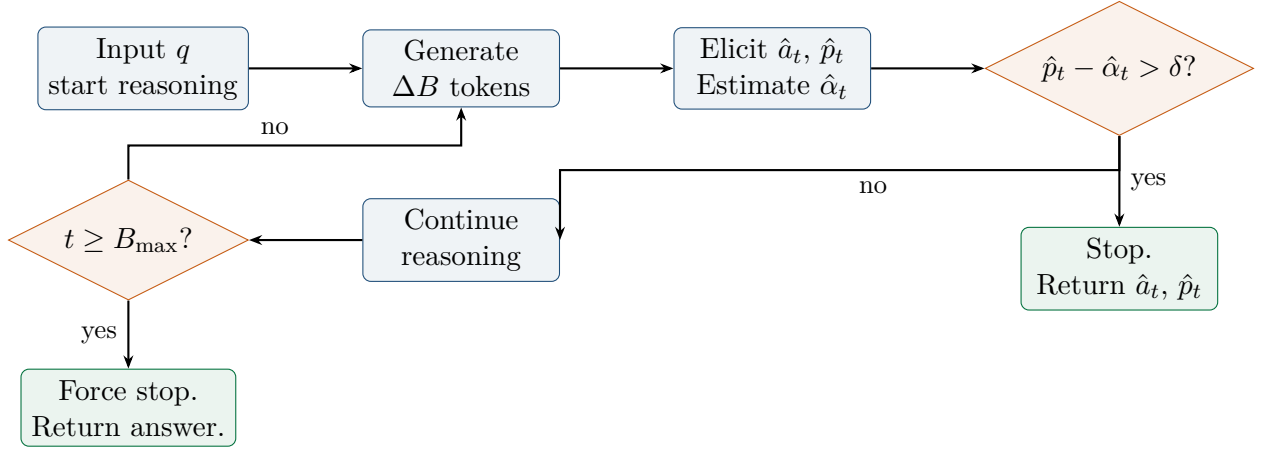

\subsection{Discussion}

\paragraph{Choice of $\delta$.}
In our experiments, OG $> 0.10$ consistently corresponds to what practitioners would consider problematic overconfidence. We recommend $\delta = 0.10$ as a starting point, which aligns with the conventional ECE threshold for ``well-calibrated'' systems. The appropriate $\delta$ is task- and deployment-dependent and should be tuned accordingly.

\paragraph{Auxiliary accuracy estimation.}
The most straightforward implementation uses self-consistency \citep{wang2022self}: generate $k$ independent continuations from the current reasoning state and use the fraction agreeing with $\hat{a}_t$ as $\hat{\alpha}_t$. With $k=5$, this adds modest compute overhead relative to the inference budget.

\paragraph{Failure cases of CABStop.}
\label{sec:cabstop_failures}
\CABStop{} assumes $\hat{\alpha}_t$ provides a meaningful signal about correctness. This assumption fails in at least two important scenarios.

First, when the model is \emph{consistently wrong across samples}: if all $k$ self-consistency samples agree on an incorrect answer, $\hat{\alpha}_t$ will be high (reflecting consistency) even though accuracy is zero. In this case, $\Delta_t = \hat{p}_t - \hat{\alpha}_t$ will be small and \CABStop{} will not halt early---precisely when it should. This failure mode is most likely on questions in the high-frequency trap categories (counting, syllogism), where the model may have a strong prior toward a specific wrong answer.

Second, when confidence is not monotone in $t$: the optimality guarantee of Proposition~6.1 requires $\hat{p}_t$ to be non-decreasing. In practice, verbalized confidence may fluctuate, especially when the model partially reconsiders an intermediate step. In this case, the stopping rule may trigger and release prematurely.

Both failure cases underscore that \CABStop{} is a heuristic whose effectiveness depends on the quality of $\hat{\alpha}_t$ and the behavioral properties of the model. Empirical validation of the algorithm on a held-out question set is a necessary step before deployment.

\paragraph{Practical recommendations.}
\label{sec:practical}
Based on our empirical observations, we offer the following guidance for practitioners choosing reasoning budgets for calibration-sensitive applications.

For tasks requiring precise discrete computation (counting, set theory, modular arithmetic, combinatorics), light reasoning budgets appear particularly risky: our results suggest they increase confidence without commensurate accuracy gains. Either no reasoning (for speed) or heavy reasoning (for accuracy) is preferable to a brief chain-of-thought that may lock in a wrong answer.

For tasks requiring multi-step logical inference (syllogisms, conditionals), the risk of early lock-in is high. Self-consistency checking---generating multiple independent chains and comparing answers---is a low-cost intervention that can improve confidence calibration without increasing single-chain budget.

Monitoring the OG metric ($\overline{\Conf} - \overline{\Acc}$) on a held-out calibration set, separately for each reasoning budget, is recommended for any deployment where calibration matters. A model showing OG $> 0.20$ at a given budget should be treated with caution, even if its accuracy is acceptable.

% ============================================================
\section{Threats to Validity}
\label{sec:threats}
% ============================================================

We report the following threats transparently. Some of these are discussed in detail in earlier sections; we consolidate them here for clarity.

\paragraph{Small sample size.}
The dataset contains 47 trap questions, yielding approximately 8--10 valid responses per trap type per model per budget. This limits the statistical power of all analyses, and particularly per-category analyses. ECE estimates are sensitive to outliers at this sample size. Confidence intervals across seeds partially mitigate this, but the standard deviations reported in Table~\ref{tab:fullresults} are large in several conditions, indicating that results should be interpreted as directional rather than precise.

\paragraph{Validity filtering and selection bias.}
As discussed in Section~\ref{sec:validity}, the 42\% validity rate may introduce systematic bias. If models are more likely to provide parseable responses when confident, our ECE estimates will be biased toward overconfidence. The direction of this bias is consistent with our main finding, meaning we cannot rule out the possibility that the observed overconfidence is partly an artifact of selection.

\paragraph{Verbalized confidence limitations.}
As discussed in Section~\ref{sec:conf_elicitation}, verbalized confidence is not equivalent to model-internal epistemic uncertainty. Results should be interpreted as measuring a behaviorally expressed property---verbalized confidence---rather than true uncertainty.

\paragraph{Model family restriction.}
We evaluate only models from the Llama-3 family. Whether \CDUR{} manifests similarly in GPT-4o, Claude-3.5, Gemini, or other families is an open empirical question. Based on our theoretical analysis, we expect the phenomenon to be present in any model with RLHF-style training, but its severity may vary substantially.

\paragraph{Prompt sensitivity.}
Budget levels are implemented via prompt engineering, which may interact with model-specific instruction-following behavior. The specific wording of budget prompts could affect both reasoning quality and confidence elicitation, and we have not conducted ablations over prompt formulations.

\paragraph{Incomplete 70B data.}
As discussed in Section~\ref{sec:70b_gap}, the absence of multi-budget data for the 70B model means that scale-related conclusions are severely limited. We do not draw conclusions about the scaling behavior of \CDUR{}.

% ============================================================
\section{Discussion}
\label{sec:discussion}
% ============================================================

\subsection{When CDUR Should Be Expected}

The \CDUR{} phenomenon is not expected to manifest equally across all task types. Based on our theoretical model and empirical results, calibration drift under reasoning is most likely for tasks that share the following properties.

\textbf{Discrete, exact-answer structure.} Problems with a unique correct answer determined by precise discrete computation (counting, modular arithmetic, combinatorics) leave little room for the model to ``almost be right.'' A wrong initial hypothesis is simply wrong, and reasoning that elaborates on it increases confidence without improving accuracy. Our results confirm that these categories dominate the wrong-and-confident distribution.

\textbf{Non-obvious traps.} Problems where a plausible-seeming wrong answer exists (e.g., anchoring, spurious pattern completion, base-rate neglect) are particularly vulnerable, because the initial hypothesis sampled by the model is likely to be the trap answer. RLHF-trained models, which are rewarded for fluent and confident responses, may be especially prone to sampling the most plausible-seeming hypothesis rather than the correct one.

\textbf{Short reasoning chains insufficient for correction.} For problems requiring multiple precise logical steps to reach the answer, short reasoning chains may correctly identify the problem type but fail to execute all necessary steps, producing a confident partial answer.

Conversely, for open-ended generation tasks, subjective evaluation tasks, and tasks with many acceptable answers, the lock-in mechanism is less relevant: there is no unique correct $\hzero$, and confidence-accuracy alignment is harder to measure.

\subsection{Implications for Inference-Time Scaling}

Recent work proposes scaling inference-time compute as a complementary axis to training-time compute \citep{snell2024scaling}. Our findings suggest a nuance: inference-time scaling improves accuracy but may worsen calibration at intermediate budget levels. A system that scales compute without monitoring calibration may present confidently wrong answers to users at the budget levels where the scaling is cheapest.

The \CABStop{} algorithm offers a concrete mechanism for calibration-aware scaling: allocate more compute when confidence and estimated accuracy agree, halt when they diverge.

\subsection{Implications for Model Evaluation}

Standard benchmarks measure accuracy. Our results suggest that calibration should be measured alongside accuracy, particularly as a function of reasoning budget. A model that achieves high accuracy at heavy reasoning but poor ECE at light reasoning may perform poorly in deployment scenarios where the full reasoning budget is not always available.

We recommend that future model evaluations report $(\Acc(B), \ECE(B))$ curves across multiple budgets, not just peak accuracy.

\subsection{Theoretical Limitations}

The Hypothesis Lock-In Model (Section~\ref{sec:mechanism}) is a stylized approximation. Real autoregressive LLMs can revise their hypothesis mid-chain, especially when prompted with explicit revision instructions. The propositions in Section~\ref{sec:theory} should be understood as characterizing a specific mechanistic regime---where the model's context is dominated by an early committed hypothesis---rather than as universal laws governing all LLM reasoning.

The commitment model also does not account for the structured attention patterns of Transformer architectures, temperature effects on hypothesis sampling, or the effect of instruction-following fine-tuning on the probability of self-correction. A more precise mechanistic account would require analysis at the level of attention weights and token probabilities, which is left to future work.

\subsection{Connections to Human Reasoning}

The \CDUR{} phenomenon has a human cognition analog: \emph{post-hoc rationalization} \citep{haidt2001emotional}. Humans often form an initial intuitive judgment and then construct reasoning that justifies it, increasing their confidence without increasing its accuracy. The commitment model in Section~\ref{sec:mechanism} formalizes this structure in the LLM setting. The parallel suggests that \CDUR{} may be a general feature of systems trained to produce justified conclusions rather than accurate ones, extending beyond LLMs to any architecture with a generate-then-justify structure.

% ============================================================
\section{Future Work}
\label{sec:future}
% ============================================================

\paragraph{Empirical replication at scale.}
The most pressing need is a larger dataset---at least 500 trap questions---to provide robust per-category calibration estimates and to verify the non-monotone ECE trajectory with statistical significance.

\paragraph{Across model families.}
Testing \CDUR{} on GPT-4o, Claude-3.5, and Gemini models would determine whether the phenomenon is universal or Llama-specific. Completing the 70B evaluation across all budget conditions is a near-term priority.

\paragraph{Training interventions.}
RLHF may be modified to reward calibrated confidence rather than expressed confidence. Calibration-aware reward models---which penalize high verbalized confidence on wrong answers---are a natural next step.

\paragraph{Better auxiliary estimators.}
\CABStop{} depends on the quality of $\hat{\alpha}_t$. Lightweight verifier models, reward models, or retrieval-augmented consistency checks could improve the accuracy estimate without substantial compute overhead, and would address the consistent-wrong-answer failure case.

\paragraph{Adaptive budgeting.}
A learned policy that maps (question, current calibration gap) to (continue/stop) would be a stronger version of \CABStop{}. This could be framed as a reinforcement learning problem with a calibration-aware reward, and would avoid the need for a hand-tuned threshold $\delta$.

\paragraph{Formal lower bounds.}
Can we prove a lower bound on the overconfidence gap under the commitment model for specific problem classes? This would provide a theoretical floor on achievable calibration and would clarify when no amount of budget tuning can resolve the overconfidence problem.

\paragraph{Per-sample calibration analysis.}
As discussed in Section~\ref{sec:scatter_discussion}, future work should report per-sample confidence-accuracy distributions in addition to aggregate ECE, to distinguish between uniform miscalibration and concentrated dangerous overconfidence.

% ============================================================
\section{Conclusion}
\label{sec:conclusion}
% ============================================================

We have introduced \textbf{Calibration Drift Under Reasoning} (\CDUR{}): the phenomenon whereby increasing reasoning budget may first improve and then worsen model calibration, producing non-monotone dynamics in the $\ECE(B)$ curve. We have provided:

\begin{enumerate}[leftmargin=*, label=(\arabic*)]
  \item A formal definition of \CDUR{} and three propositions characterizing it under a probabilistic Hypothesis Lock-In Model, establishing theoretical grounding for the phenomenon under an explicit mechanistic abstraction.

  \item Empirical evidence from Llama-3.1-8B on 47 reasoning-trap questions spanning 21 cognitive failure modes, showing non-monotone ECE dynamics and persistent overconfidence gaps exceeding 0.25 across all budget levels. We report limited Llama-3.3-70B results (no-reasoning only) and explicitly acknowledge that \CDUR{} dynamics at this scale remain unconfirmed.

  \item The \CABStop{} algorithm---a calibration-aware stopping rule grounded in optimal stopping theory---which halts reasoning when the confidence-accuracy gap exceeds a threshold. We also characterize its failure cases and discuss practical guidance for threshold selection.
\end{enumerate}

We have been explicit about the methodological limitations of this study: the small dataset, the 42\% validity rate and its potential for selection bias, the imprecision of verbalized confidence as an uncertainty signal, and the incomplete multi-scale evaluation. These limitations do not negate the value of the theoretical framework or the empirical observations for the 8B model, but they do call for the results to be treated as preliminary evidence motivating further investigation rather than as settled empirical findings.

The central message is not that more reasoning is always worse for calibration, but that the relationship between reasoning depth and calibration quality is not monotone, and that this non-monotonicity can be dangerous in practice. As inference-time scaling becomes a mainstream technique, calibration monitoring---ideally combined with adaptive stopping rules---deserves attention as a first-class concern alongside accuracy optimization.

\vspace{16pt}
\hrule
\vspace{10pt}

% ============================================================
% Bibliography
% ============================================================

% ============================================================
\appendix
% ============================================================

\section{Proof Details}
\label{app:proofs}

\subsection{Formal Consistency Score}

We define the consistency score $\mathcal{C}$ used in Proposition~3.4 as follows. Let $\mathcal{T}$ be the vocabulary and $V_{\hzero} \subset \mathcal{T}$ be the set of tokens that reinforce hypothesis $\hzero$ (e.g., tokens that share lexical content with $\hzero$ or that appear in reasoning chains that conclude with $\hzero$, as measured by a reference corpus). Then:
\[
  \mathcal{C}(R_1,\ldots,R_t; \hzero) = \frac{1}{t}\sum_{i=1}^{t} \mathbf{1}[R_i \in V_{\hzero}].
\]
This is a running fraction, and $\E[\mathcal{C}] = \Prob[R_i \in V_{\hzero} \given \hzero] =: \rho(\hzero) \in (0,1)$ by assumption.

Under the commitment model, $\Prob[R_i \in V_{\hzero} \given \hzero, R_1,\ldots,R_{i-1}]$ is non-decreasing in $i$ (since the context increasingly supports $\hzero$), so $\mathcal{C}$ is a submartingale with positive drift for all $t \geq 1$. Hence $\E[\mathcal{C}(R_1,\ldots,R_t;\hzero)]$ is non-decreasing in $t$.

\subsection{Discussion of Assumption Robustness}

The key assumption is that $P_\theta(R_i \in V_{\hzero} \given \hzero, R_{<i})$ is non-decreasing in $i$. This holds under self-attention architectures when:
\begin{enumerate}
  \item $\hzero$ appears in the first few tokens of the reasoning chain and thus in the attention window.
  \item The model has been trained to produce coherent, self-consistent outputs (as encouraged by RLHF).
\end{enumerate}
Both conditions hold approximately for current instruction-tuned LLMs. We do not claim they hold universally, and the assumption may be violated when the model encounters strong contradictory evidence mid-chain or when explicit revision instructions are provided.

\section{Dataset: Trap Question Examples}
\label{app:examples}

We provide representative examples from four high-frequency trap categories.

\paragraph{Counting (highest frequency).}
\textit{Question:} ``A frog climbs 3 meters up a 10-meter wall each day and slides back 2 meters each night. How many days to reach the top?''
\textit{Expected:} 8. \textit{Common wrong answer:} 10. \textit{Trap:} On day 8, the frog reaches 10m during the day before sliding back; students often compute 10/1 = 10 naively.

\paragraph{Set theory.}
\textit{Question:} ``In a class of 30, 18 play football, 15 play cricket, and 5 play both. How many play neither?''
\textit{Expected:} 2. \textit{Common wrong answer:} 3. \textit{Trap:} Inclusion-exclusion requires $|F \cup C| = 18 + 15 - 5 = 28$; neither = $30 - 28 = 2$.

\paragraph{Syllogism.}
\textit{Question:} ``All A are B. All B are C. Is it true that all C are A?''
\textit{Expected:} No. \textit{Common wrong answer:} Yes. \textit{Trap:} The converse of a universal statement does not follow.

\paragraph{Probability.}
\textit{Question:} ``A box has 2 red and 2 blue balls. You draw 2 without replacement. What is the probability both are red?''
\textit{Expected:} $1/6$. \textit{Common wrong answer:} $1/4$. \textit{Trap:} The draws are dependent; $P = \frac{2}{4} \cdot \frac{1}{3} = \frac{1}{6}$.

\section{Experimental Logs (Summary)}
\label{app:logs}

Table~\ref{tab:fullresults} reproduces the full summary statistics from the experimental runs. Standard deviations are computed across 3 seeds; the 70B model was not evaluated at multi-budget conditions, and those entries are absent.

\begin{table}[h]
\centering
\small
\caption{Full results: ECE (mean $\pm$ std across 3 seeds), overconfidence gap, and accuracy for trap questions. The large standard deviation at 8B medium budget reflects high seed-to-seed variability and should be interpreted cautiously. 70B results are available only at the no-reasoning condition.}
\label{tab:fullresults}
\begin{tabular}{@{}llccc@{}}
\toprule
\textbf{Model} & \textbf{Budget} & \textbf{ECE $\pm$ std} & \textbf{OG (conf$-$acc)} & \textbf{Accuracy} \\
\midrule
Llama-3.1-8B  & none   & $0.0436 \pm 0.0154$ & $+0.493$ & $0.461$ \\
Llama-3.1-8B  & light  & $0.1040 \pm 0.0339$ & $+0.249$ & $0.732$ \\
Llama-3.1-8B  & medium & $0.0496 \pm 0.0492$ & $+0.336$ & $0.653$ \\
Llama-3.1-8B  & heavy  & $0.0145 \pm 0.0049$ & $+0.245$ & $0.739$ \\
\midrule
Llama-3.3-70B & none   & $0.0352 \pm 0.0264$ & $+0.155$ & $0.825$ \\
Llama-3.3-70B & light  & \multicolumn{3}{c}{\textit{not evaluated}} \\
Llama-3.3-70B & medium & \multicolumn{3}{c}{\textit{not evaluated}} \\
Llama-3.3-70B & heavy  & \multicolumn{3}{c}{\textit{not evaluated}} \\
\bottomrule
\end{tabular}
\end{table}

\end{document}